\documentclass[letterpaper,journal]{IEEEtran}
\usepackage{amsmath,amsfonts}
\usepackage{algorithmic}
\usepackage{algorithm}
\usepackage{array}
\usepackage[caption=false,font=normalsize,labelfont=sf,textfont=sf]{subfig}
\usepackage{textcomp}
\usepackage{stfloats}
\usepackage{url}
\usepackage{verbatim}
\usepackage{graphicx}
\usepackage{cite}
\usepackage{siunitx}
\usepackage{orcidlink}
\newcommand{\ace}{\textit{Ace}}

\usepackage{svg}
\usepackage{bm}
\usepackage{adjustbox}
\usepackage{graphicx}
\usepackage[utf8]{inputenc}
\DeclareUnicodeCharacter{2212}{\textminus}

\begin{document}

\title{\LARGE \bf
Hardware Design for Table Tennis Robot \\Capable of Beating Professional Players}

\author{Nobuhiko~Mukai$^*$, 
        Pavel~Adodin$^*$,
        Stefan~Heusser$^*$,
        Alexander~Sigrist$^*$,
        Divij~Grover$^*$,
        Guillem~Torrente,
        Farshad~Khadivar,
        Takekazu~Kakinuma,
        and~Peter~D\"urr%
        % <-this % stops a space

\thanks{Nobuhiko Mukai, Pavel Adodin and Guillem Torrente are with Sony AI in Tokyo, Japan.
Stefan Heusser, Alexander Sigrist, Divij Grover, Farshad Khadivar and Peter D\"urr are with Sony AI in Z\"urich, Switzerland.
Takekazu Kakinuma is with Sony Global Manufacturing \& Operations in Tokyo, Japan.}%
\thanks{*These authors contributed equally to this work.}%
\thanks{The supplementary video is available in the ancillary files. %
%(\href{anc/supplemental_video.mp4}{anc/supplemental\_video.mp4})
}%
}

\maketitle

\begin{abstract}
This paper focuses on the hardware specifications required for a table tennis robot to beat professional players.
After analyzing the motions of elite players, we defined target specifications for the workspace, payload, external-force resistance, physical performance, serve capability, and end-effector accuracy. 
Based on these specifications, we developed \ace, a custom 8-DoF robot. The mechanical structure was improved through topology optimization to minimize mass while preserving stiffness. 
Motor and gearbox selection was optimized using an inverse-dynamics torque model. 
Low-order per-joint dynamics models with delay compensation were identified and integrated into simulation to enable the use of an RL control policy. 
Experiments demonstrated repeated full-stroke swings with a cycle time of $\mathbf{0.8~\si{s}}$ and a peak racket-center velocity of $\mathbf{22~\si{m/s}}$. 
The robot successfully defeated multiple professional players. 
\end{abstract}


%%%%%%%%%%%%%%%%%%%%%%%%%%%%%%%%%%%%%%%%%%%%%%%%%%%%%%%%%%%%%%%%%%%%%%%%%%%%%%%%
%%%%%%%%%%%%%%%%%%%%%%%%%%%%%%%
\section{Introduction}
%%%%%%%%%%%%%%%%%%%%%%%%%%%%%%%
%\IEEEPARstart{T}{able} 
Table
tennis is a sport that requires fast reaction times as the ball velocity can exceed $20~\si{m/s}$, the spin (i.e., the angular velocity of the ball) can reach
$1000~\si{rad/s}$, and the flight time of the ball between shots is often less than $0.25~\si{s}$ \cite{TT_statics}. For a robot to compete on equal terms with elite players, high-speed and high-precision strokes required to generate high-speed and high-spin balls must be repeatedly executed within a short time frame. This requires precise tracking of the racket reference trajectory generated by controllers such as AI agents with minimal delay, based on information about the ball and the opponent obtained from the perception system.

Previous research on table tennis robots has often centered on commercially available industrial robots \cite{kyohei2020ping, d2024achieving}. However, no industrial robot has been commercially released that possesses the performance—in terms of motion acceleration, velocity, and workspace—to compete against professional players. Furthermore, to the best of the authors' knowledge, no table tennis robot has been custom-designed with the goal of actually competing in matches and defeating professional players.

\begin{figure}[!t]
\centering
\includegraphics[width=8cm]{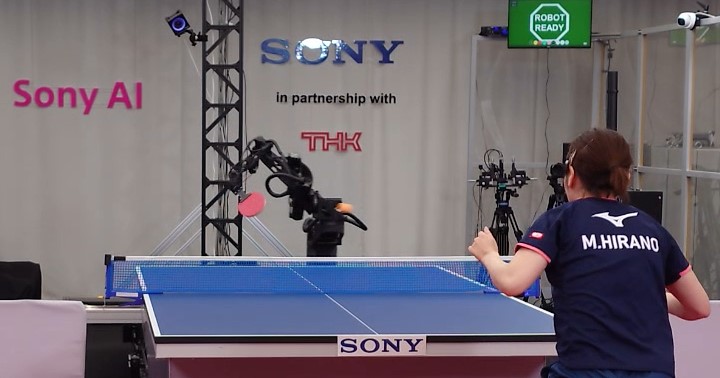}
\caption{\textbf{
Matches against professional table tennis players.} 
In the spring of 2026, \ace{} competed against 9 professional players and defeated 8 players, including Miu Hirano (Olympic silver medalist at Tokyo 2020 and Paris 2024, highest world ranking of No.5) (see the supplementary video)
\cite{sonyai2026blog}.
}
\label{fig:play}
\end{figure}

\begin{figure}[!t]
\centering
\includegraphics[width=8cm]{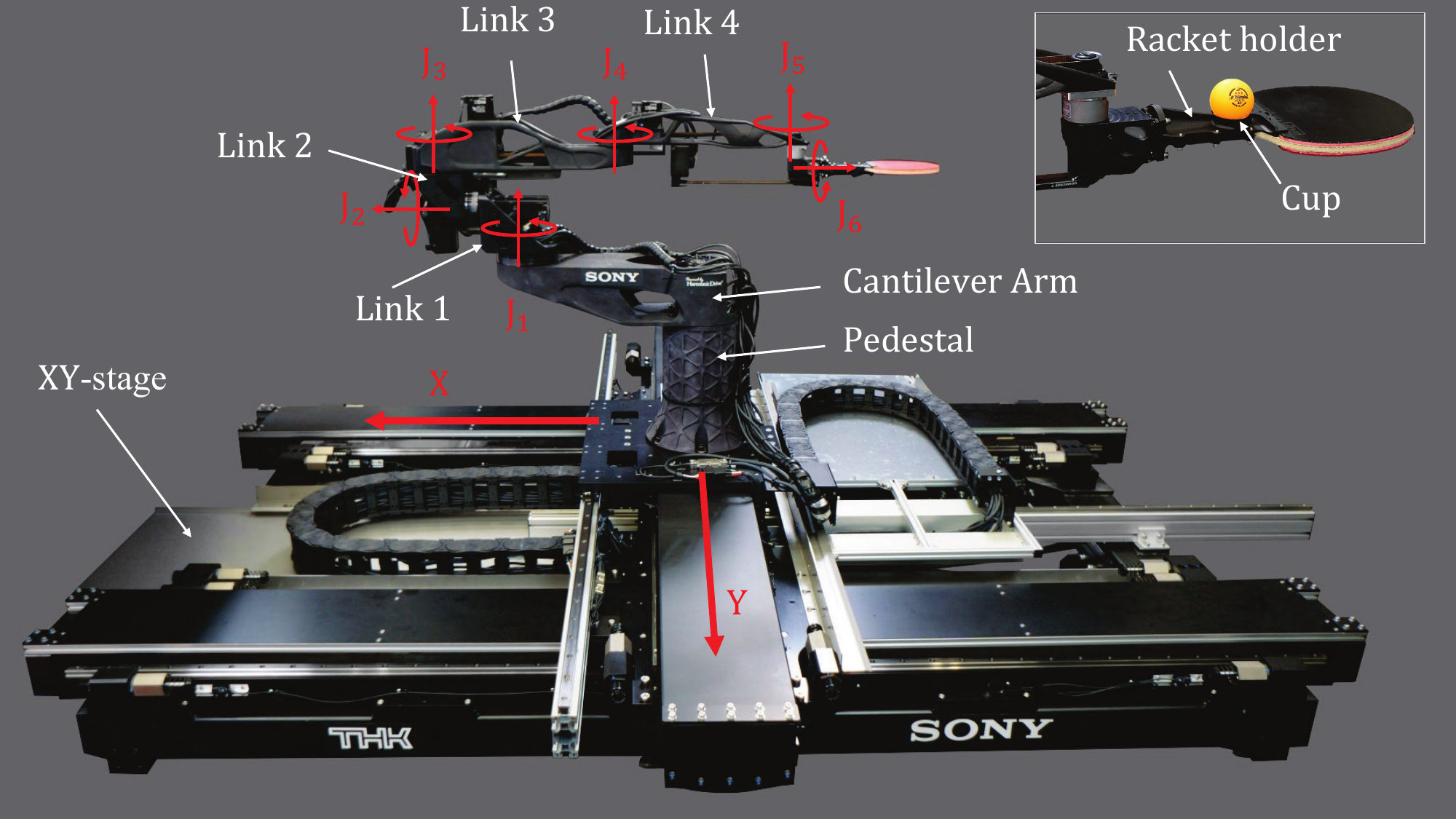}
\caption{\textbf{Table tennis robot \ace{}.} 
It is an 8-DoF robot that mounts a 6-DoF arm on an XY-stage. It has an end effector
equipped with a racket and a cup to hold the ball, facilitating one-handed
serves  in conjunction with an external ball feeder (see the supplementary video for the movement of each axis and the one-handed serve). 
}
\label{fig:outline_XY}
\end{figure}

We designed \ace, a novel robot to enable matches against professional players (Fig.~\ref{fig:play}, 
Fig.~\ref{fig:outline_XY}) \cite{duerr2026Ace}.
This robotic platform complies with the International Table Tennis Federation (ITTF) rules \cite{ittf_statutes}, with the exception of safety-related modifications, such as the prohibition of players entering the opponent's side of the court and side changes.

%%%%%%%%%%%%%%%%%%%%%%%%%%%%%%%%%%%%%%%%%%%%%%%%%%%%%%%%%%%%%%%%%%%%%%%%%%%%%%%%%%%%%%%%
\section{Setting Target Specifications }\label{sec:target_specs}
\begin{figure}[!t]
\vspace{6pt} 
\centering
\includegraphics[width=7.5cm]{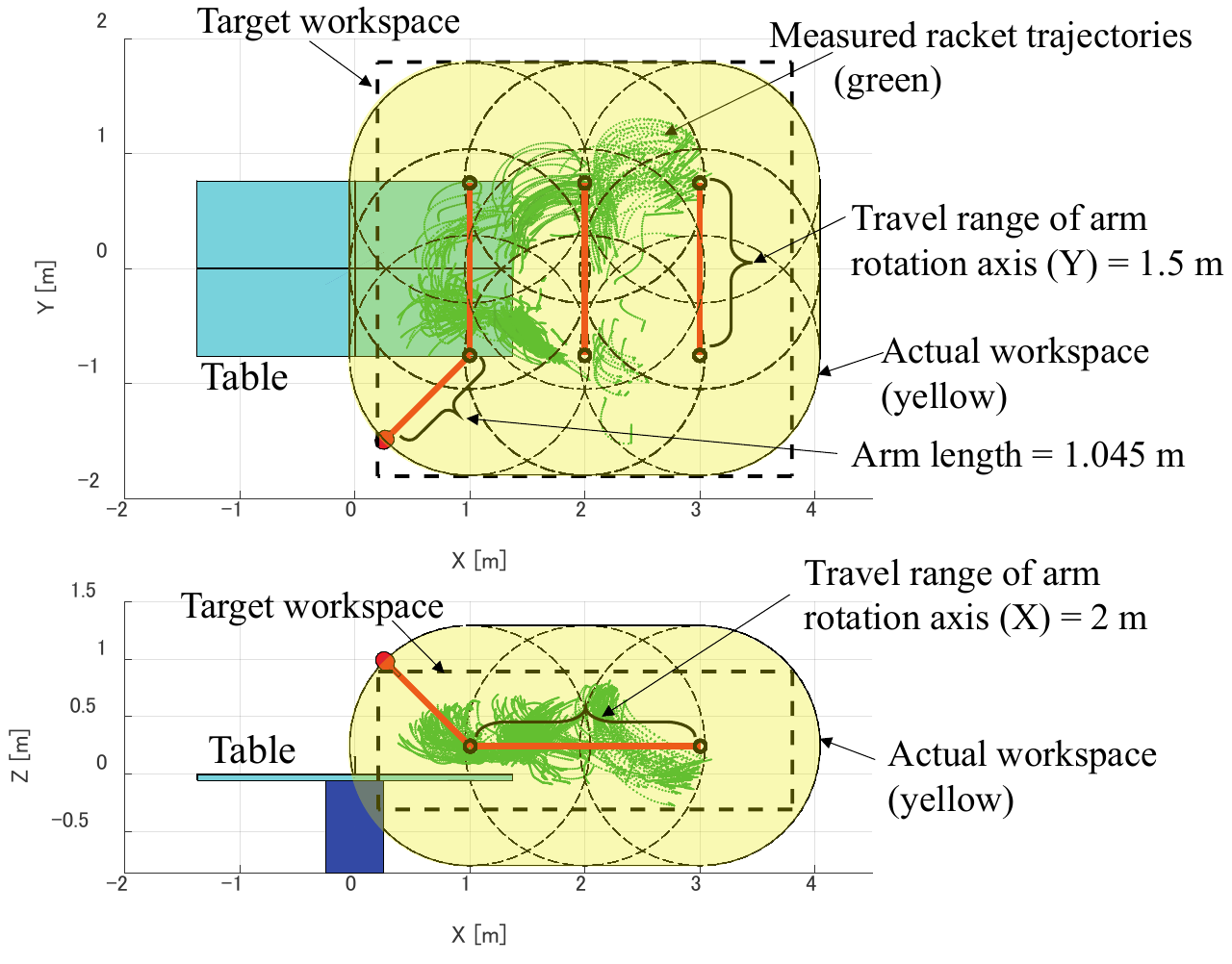}
\caption{\textbf{Determined workspace based on measured racket trajectories.} 
Travel range of arm rotation axis, arm length, and the workspace were determined based on the movement of the racket centers for various strokes during serves and rallies by advanced players.
}
\label{fig:workspace}
\end{figure}

%%%%%%%%%%%%%%%%%%%%%%%%%%%%%%%%%%%%%%%%%%%%%%%%%%%%%%%%%%%%%%%%%%%%%%%%%%%%%%%%%%%%%%%%
We recorded two advanced table tennis players with university league experience while they repeatedly performed example motions for various strokes during serves and rallies (backhand/forehand, topspin/push/block/flick, etc.). 
Based on the three-dimensional trajectory of their racket centers, measured using an optical motion capture system, the target workspace was determined to be 3.6 m x 3.6 m horizontally (covering half the table), and $0.3~\si{m}$ below and $0.9~\si{m}$ above the tabletop (Fig.~\ref{fig:workspace}).

\begin{figure}[!tt]
\centering
\includegraphics[width=7.5cm]{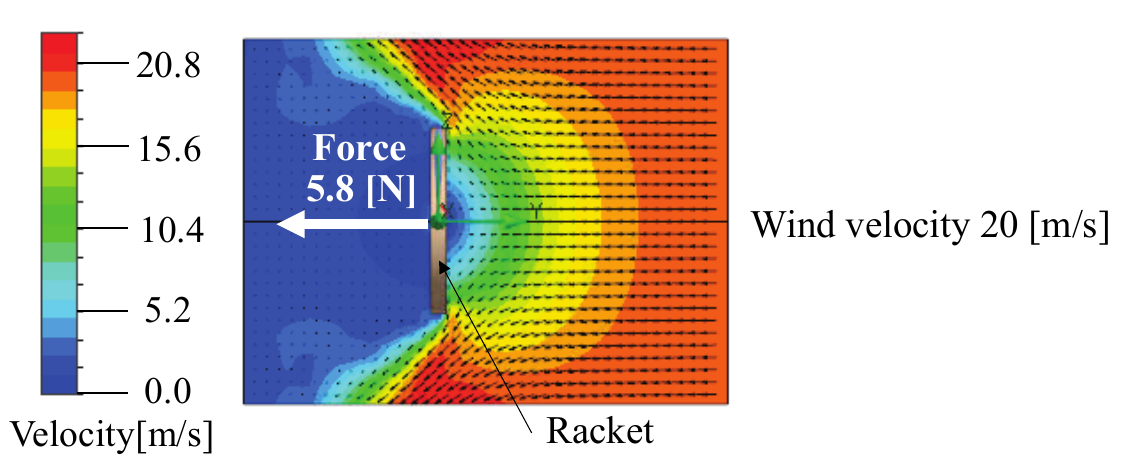}
\caption{\textbf{Airflow simulation of a racket.} 
The force acting on the end-effector holding a racket was obtained as $5.8~\si{N}$ at a wind velocity of $20~\si{m/s}$.
}
\label{fig:air_flow}
\end{figure}

The only payload required for a table tennis robot is the racket (around $180~\si{g}$). However, the air resistance acting on the racket during the stroke is an external force at the end-effector, which in past high-speed motion experiments using a commercial industrial robot caused alarm stops due to insufficient torque, depending on the racket angle. Our simulation estimated external force of approximately $5.8~\si{N}$ on the racket with $20~\si{m/s}$ wind velocity (Fig.~\ref{fig:air_flow}). Despite the modest magnitude of this force, a torque margin capable of resisting at least $6~\si{N}$ of external force must be secured at the end-effector during high-speed motion.


Previous studies measuring matches between top-tier table tennis players have reported that 
the mean waiting time (until a player hits the ball after the opposing player has hit it) 
was $0.66\pm0.19~\si{s}$, and that  
the shot interval time per player (i.e., the time between a player hitting the ball and that same player hitting it again) during rallies was estimated to range from $1.0~\si{s}$ to $1.8~\si{s}$ \cite{Yoshida1991}.
We also measured the shot interval time between two players who play in a Japanese professional league. During warm-up rallies (standing near the table without changing position and repeating moderate forehand drives), the mean shot interval time was $0.79~\si{s}$ with a minimum of $0.77~\si{s}$. During match rallies (including position changes and various racket swing motions), the mean shot interval time was $1.01~\si{s}$ with a minimum of $0.67~\si{s}$ (see the supplementary video). 

Based on these measurements, we determined the physical performance targets for our robot as a full-stroke swing cycle time of $0.8~\si{s}$ with an end-effector linear velocity of $20~\si{m/s}$ matching the racket velocity in smash shots by elite players \cite{TT_statics}.

According to ITTF rules, a one-handed serve is permitted at the referee's discretion as an exception for players who cannot use both hands due to a physical disability. 
Consequently, the development of a mechanism for the one-handed serve was determined to be appropriate.  
To ensure seamless gameplay, the incorporation of an automatic ball-feeding mechanism was also deemed essential.

In order to operate a robot with both high-speed and high-precision using Reinforcement Learning (RL) control policies, it is not only essential that the low-level position control system exhibits minimal tracking error, but it is also necessary to identify highly accurate and predictable dynamic models to facilitate the training of high-level RL control policies in simulation.

We established our target for the positional and rotational accuracy needed at the end effector assuming that \ace{} 
hits the ball precisely at the racket's center point.
A circle with radius of $7~\si{cm}$ can be inscribed inside the racket blade, providing an upper bound for the desired positional accuracy. 
However, considering the effect of the racket angular velocity (which can reach  $20~\si{rad/s}$) on the post-contact ball linear velocity, the target positional accuracy was ultimately set to $0.4~\si{cm}$, which is $10~\si{\%}$ of the ball’s diameter.

For the rotational accuracy, the desired ball landing accuracy on the opponent's side of the table was assumed to be $10~\si{cm}$. 
Subsequently, the angle at which a $10~\si{cm}$ positional deviation occurs when a robot hits a ball from a position $2~\si{m}$ behind the net and $0.3~\si{m}$ above the table surface, and the ball lands at the edge of the opponent's court ($1.37~\si{m}$ from the net) was calculated as follows:
 $\theta=\arctan(0.3/(2+1.37-0.1)) - \arctan(0.3/(2+1.37))\approx0.003~\si{rad}$.
Based on the above, we compiled the target specifications in Table~\ref{tab:target_spec}.

\begin{table}[!t]
\vspace{6pt} 

\centering
\caption{Target specifications for a competitive table tennis robot}
\label{tab:target_spec}
\begin{tabular}{|p{1.7cm}|p{6cm}|}
\hline
\textbf{Items} & \textbf{Target to be achieved} \\
\hline
\hline
Workspace & 
Horizontal: 3.6 m × 3.6 m (covering half the table)\par Vertical: -0.3 m to +0.9 m from the tabletop \\

\hline
Payload & More than 180 g (racket mass) \\
\hline
Resistance to\par external forces& More than 6 N at the end-effector \\
\hline
Physical\par performance & Full-stroke swing cycle time of 0.8 s with an end-effector linear velocity of 20 m/s \\
\hline
Serve & Incorporating mechanism for a one-handed serve with an automatic ball-feeding function \\
\hline
Modeling\par accuracy & Positional error at the end-effector: under 0.4 cm\par  
Rotational error at the end-effector: under 0.003 rad \\
\hline
\end{tabular}
\end{table}

%%%%%%%%%%%%%%%%%%%%%%%%%%%%%%%%%%%%%%%%%%%%%%%%%%%%%%%%%%%%%%%%%%%%%%%%%%%%%%%%%%%%%%%%
\section{Design of the Basic Configuration}
%%%%%%%%%%%%%%%%%%%%%%%%%%%%%%%%%%%%%%%%%%%%%%%%%%%%%%%%%%%%%%%%%%%%%%%%%%%%%%%%%%%%%%%%

To achieve the workspace and agility required for professional-level table tennis (shown in Section \ref{sec:target_specs}), we developed the table tennis robot \ace{},
a custom robot platform featuring eight degrees of freedom (two prismatic and six revolute joints; Fig.~\ref{fig:outline_XY}), which was determined as the minimum number necessary to execute competitive shots: three for the position of the racket, two for its orientation, and three for the velocity vector and magnitude of the shot.  
The 6-DoF arm was placed on the cantilever-shaped base to facilitate the use of the entire court as a workspace, with J1 of the robot arm able to be positioned above the tabletop. This allowed the racket to reach the net without the need to further increase the length of moving links, thus keeping the robot as lightweight as possible. The arm is moved by a 2-axis linear stage
manufactured by THK, 
referred to as the XY-stage.
The total arm length was set to satisfy all conditions of maximum end-effector velocity, cycle time and overall workspace of the robot (Fig.~\ref{fig:workspace}).
Joints J3, J4 and J5 have their rotational axes parallel, generating maximum end-effector velocity in a single ``work plane", while joint J6 adjusts the racket inclination angle according to the ball trajectory. By using joints J1 and J2 to rotate that plane, the robot is able to generate the required end-effector velocity throughout 3D space.

The end effector is composed of a racket in a racket holder with a small cup %(Fig.~\ref{fig:racket_holder}) 
to hold and toss the ball during the serve. The racket holder is able to hold any standard ITTF-approved racket.
The symmetrical structure of \ace{} 
eliminates the distinction between forehand and backhand movements, enabling the RL control policies to efficiently learn how to control it by augmenting the training data along the plane of symmetry.
In the home position, the starting posture of the game, the racket points towards the side opposite from the table. This enables the robot to instantly move to either side of the table while accelerating efficiently, thus reducing the time needed for the robot to hit the ball with maximum velocity.
We have chosen lightweight Harmonic Drive reduction gears for all rotational axes to provide the required torque and stiffness. In addition, by using timing belt transmissions on the input side of the reduction gear where possible (namely, in joints J1, J4, J5 and J6), we placed the motors closer to the rotational axes and thus reduced the inertia of the robot's moving parts, while keeping the stiffness and eigenfrequency of the system high.

\begin{figure}[!t]
\vspace{6pt} 
\centering
\includegraphics[width=8cm]{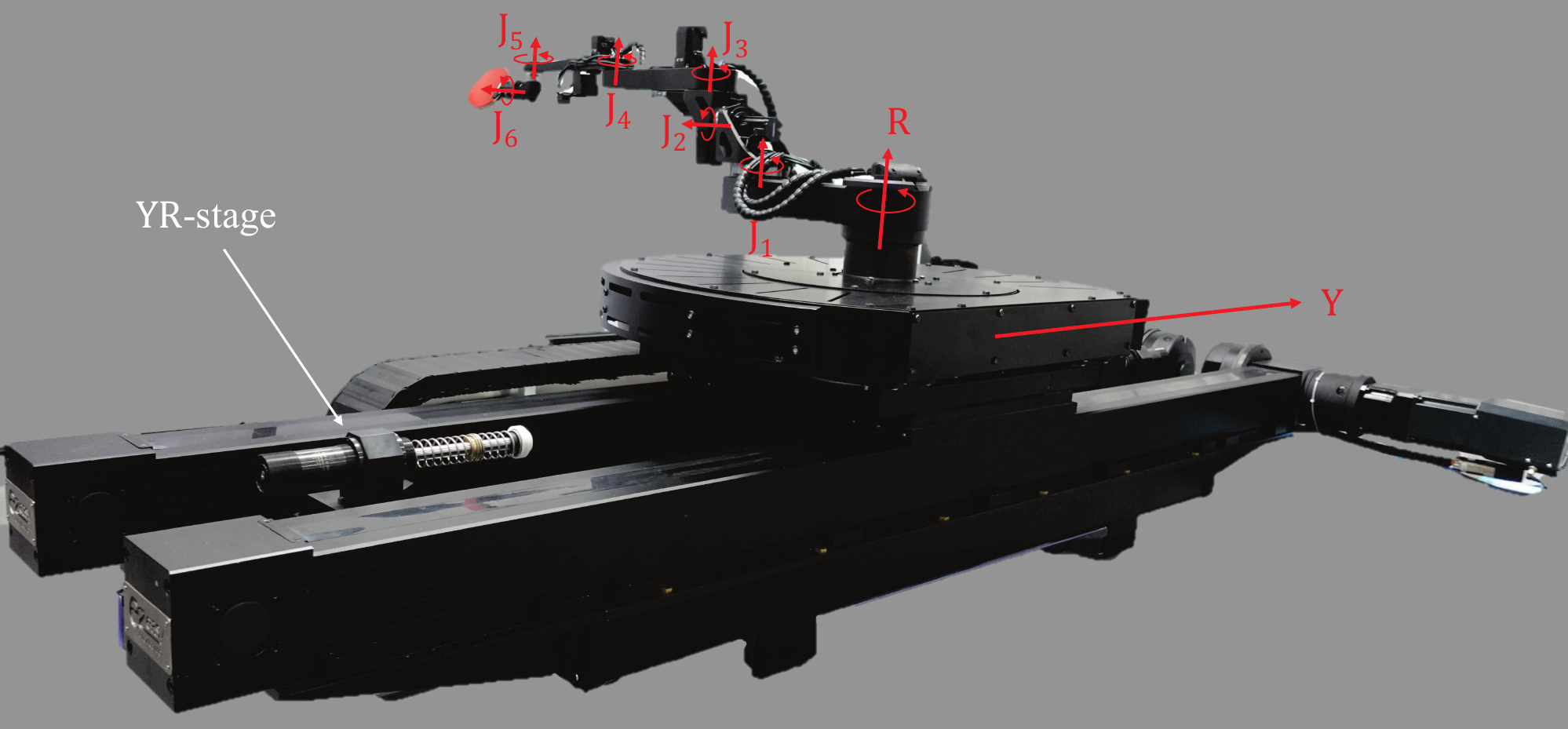}
\caption{\textbf{YR-stage configuration of \ace.}
By replacing the XY-stage with a YR-stage, the game can be played even in a small laboratory that does not meet ITTF regulations.
}
\label{fig:outline_YR}
\end{figure}

In parallel to the primary XY-stage configuration setup, an alternative base configuration was developed and deployed to investigate the kinematic advantages of a translation-rotation setup. 
In this alternative system, the 6-DoF arm remains identical, but the dual axis linear base is replaced with a single translation axis and an additional rotational axis, referred to as the YR-stage (Fig.~\ref{fig:outline_YR}).
The primary objective for the YR-stage was to continue development at our Zurich office, where workspace is limited, as it was designed to have a smaller overall footprint and working envelope. In this setup, translation along the X axis is instead achieved by combining the motion of the rotational (R) axis and the linear (Y) axis. 
One advantage of having the R axis is that it can generate even higher end-effector (racket) velocity by compounding the base (R axis) rotational velocity with the velocities of the robot joints (J1, J3, J4 and J5). However, since the R axis is parallel to J1, the angular acceleration of that is coupled with J1, and in many cases with J3, J4 and J5, resulting in higher torque required for the same motion, when compared to the linear stage (XY) configuration. 

In consideration of the larger available court area at the Tokyo office, the XY-stage setup was utilized for matches against professional players.

%%%%%%%%%%%%%%%%%%%%%%%%%%%%%%%%%%%%%%%%%%%%%%%%%%%%%%%%%%%%%%%%%%%%%%%%%%%%%%%%%%%%%%%%
\section{Mass minimization through topology optimization} \label{s:topopt}
%%%%%%%%%%%%%%%%%%%%%%%%%%%%%%%%%%%%%%%%%%%%%%%%%%%%%%%%%%%%%%%%%%%%%%%%%%%%%%%%%%%%%%%%

High-speed robotic manipulators demand a balance between structural stiffness and mass. 
Excess mass in the arm links increases inertial loads, motor torque requirements, and energy consumption, while insufficient stiffness leads to undesirable deflections and vibrations that degrade accuracy and predictability. 
Topology optimization provides a systematic, computational approach to redistribute material within a defined design space, yielding structures that are lighter while satisfying prescribed structural constraints. Combined with additive manufacturing (AM), it also enables complex, organic geometries that would otherwise be difficult to realize with conventional machining. Recent robot-arm studies have optimized link topologies under motion-dependent inertial loads, combined modal analysis with topology optimization to jointly minimize mass and vibration, and performed topology optimization on full assembly models to achieve more effective lightweight design of multi-DoF robots \cite{Wu2025,Alshihabi04032025,Sha2020}. Here, we apply those ideas to our robot and carry it through AM-aware redesign.

We first performed a baseline structural analysis of the full robot, then carried out local optimization of selected components, and finally reinterpreted the results into manufacturable CAD models. Structural loads were derived from the recursive Newton--Euler algorithm (RNEA) at worst-case robot configurations using the maximum joint velocities and accelerations. Each part was divided into design and non-design regions so that functional interfaces remained unchanged during optimization. The baseline study showed that mass reduction should focus on the distal structural parts, with Link~4 and Link~3 as the main candidates and the cantilever arm and pedestal as additional targets. Links~1 and~2 were not pursued further because they were already close to their structural limits.

For component-level optimization, we used the level-set optimizer in Altair OptiStruct \cite{AltairOptiStruct2025} to minimize mass subject to lower-bound constraints on the natural frequencies. Each study was carried out on an isolated component design space while preserving the non-design space, allowing the result to be transferred efficiently into the final CAD model. Component-specific frequency constraints were selected for each part to balance mass reduction and dynamic performance, while stiffness, peak stresses, and fatigue life were evaluated separately in the subsequent finite-element verification.

\begin{figure}[!t]
\vspace{6pt} 
\centering
\includegraphics[width=8.5cm]{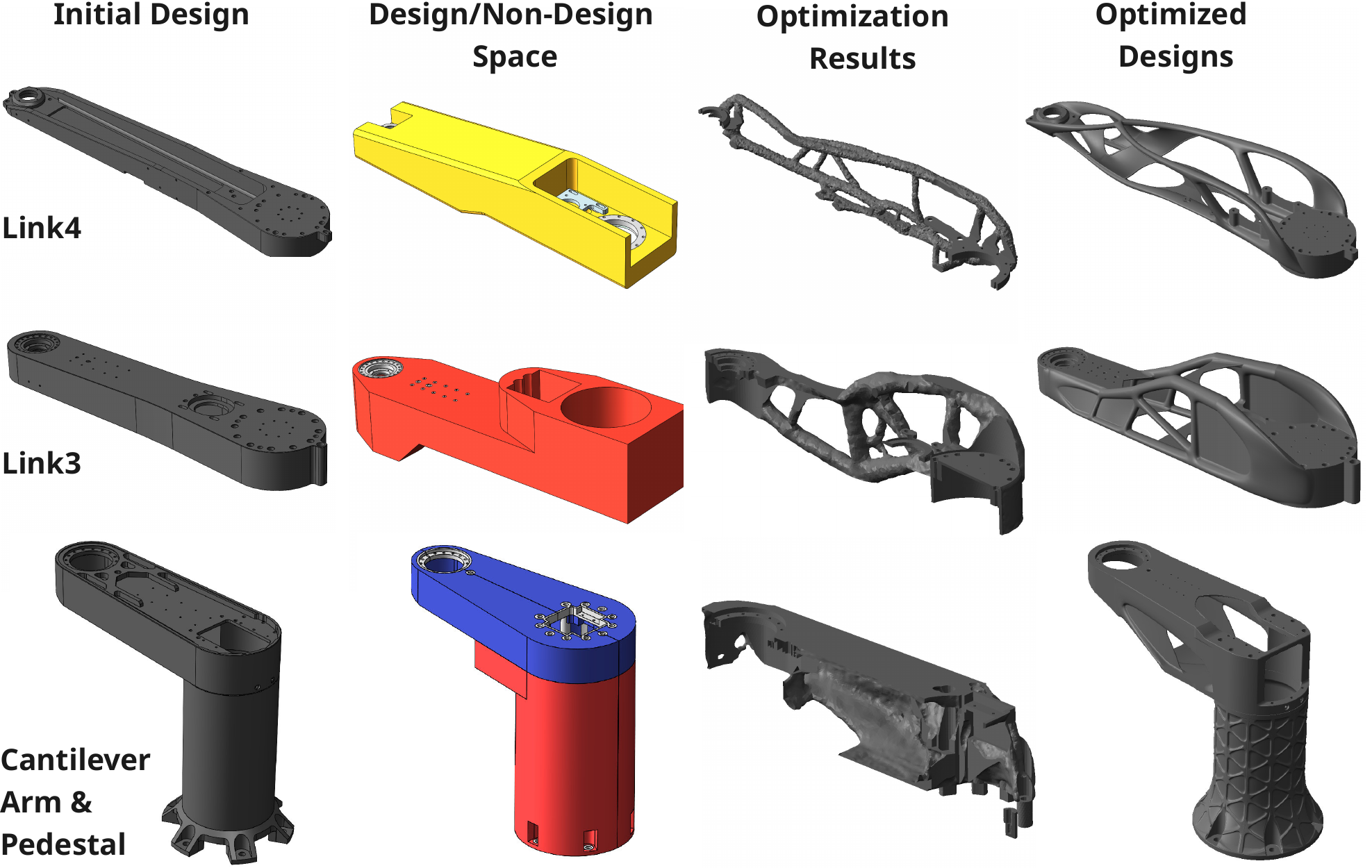}
\caption{\textbf{Topology-optimization design progression for the main structural components.}
The columns show the initial design, the optimization setup with design space shown in red, blue, and yellow and non-design space shown in gray, the raw topology-optimization result, and the final AM-ready design.}
\label{fig:topology_optimization_result}
\end{figure}

During the design process, the selected topology-optimized geometries were converted into manufacturable designs by applying practical AM design rules, including minimum wall thicknesses, machining allowances at functional interfaces, and appropriate build orientations.
The optimized links were manufactured in Scalmalloy using laser powder bed fusion (LPBF). The intermediate and final outcomes are shown in Fig.~\ref{fig:topology_optimization_result}. The final designs were verified by finite-element analysis under worst-case loading conditions derived from the RNEA study. For Link~4, the optimized design reduced mass from $0.84~\si{kg}$ to $0.60~\si{kg}$ while satisfying the structural requirements. The same workflow was applied to Link~3, the cantilever arm, and the pedestal. Mass reductions across all topology-optimized components are summarized in Table~\ref{tab:to_mass}.

\begin{table}[!t]
\renewcommand{\arraystretch}{1.1}
\centering
\caption{Mass reduction summary for topology-optimized components}
\label{tab:to_mass}
\begin{tabular}{lccc}
\hline
Component & Baseline (kg) & Optimized (kg) & Reduction (\%) \\
\hline
Link 4         & 0.84 & 0.60 & 28.57 \\
Link 3         & 2.27 & 1.70 & 25.11 \\
Cantilever Arm & 6.16 & 4.30 & 30.19 \\
Pedestal       & 6.23 & 4.00 & 35.79 \\
\hline
Total          & 15.50 & 10.60 & 31.6 \\
\hline
\end{tabular}
\end{table}

%%%%%%%%%%%%%%%%%%%%%%%%%%%%%%%%%%%%%%%%%%%%%%%%%%%%%%%%%%%%%%%%%%%%%%%%%%%%%%%%%%%%%%%%
\section{Optimization of Motor and Gearbox Selection Using the Inverse Dynamics Torque Model}
%%%%%%%%%%%%%%%%%%%%%%%%%%%%%%%%%%%%%%%%%%%%%%%%%%%%%%%%%%%%%%%%%%%%%%%%%%%%%%%%%%%%%%%%

To optimize the selection of motors and gearboxes for the robot actuators, a study was conducted using an inverse dynamics model of the robot. This process requires analysis using a model since there is a tradeoff when selecting the best actuator configuration for the task; while larger gearboxes and motors come with higher peak torque ratings, they add mass and inertia to the robot arm which also causes the required joint torque to increase for the same motion. These two factors have opposite effects on the ability to execute highly dynamic motions and the optimal choice depends on the mass properties of the structural links of the robot.

\begin{figure}[!t]
\vspace{6pt} 
    \centering
    \includegraphics[width=8.5cm]{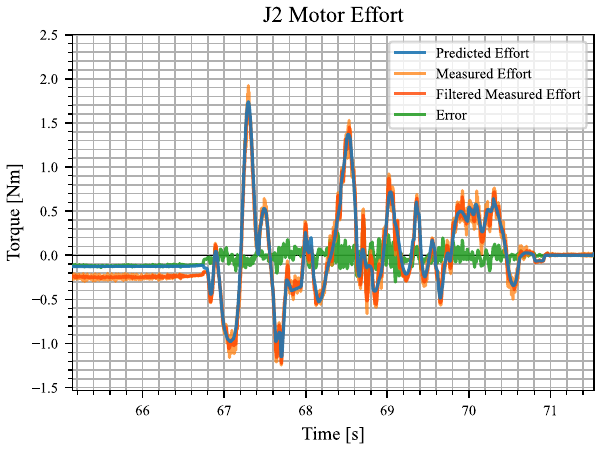}
    \caption{Plot illustrating the torque of the J2 motor over time during stroke motions. The measured torque from the motor current is shown in yellow, the filtered version of this signal is shown in red, the estimate of the motor torque from the inverse dynamics model is indicated in blue, and the error in the estimate is shown in green.}
    \label{fig:time-domain_motor_torque}
\end{figure}

Following the topology optimization of the robot links described in section \ref{s:topopt}, a re-dimensioning of the motor actuators was performed to capitalize on the inertia reductions achieved in the links.

 The inverse dynamics model predicts the torque margin for each motor and gearbox of the robot given a desired joint trajectory, $q(t)$, and an actuator configuration which specifies the exact motors and gearboxes installed at each joint. This method enables various hypothetical joint actuator configurations to be evaluated in simulation and compared before construction. The inverse dynamics model is based upon the computationally efficient RNEA implementation from Pinocchio \cite{carpentier2019pinocchio, pinocchioweb} with a bespoke transmission model on top to capture the inertia and friction of the transmission components. Fig.~\ref{fig:time-domain_motor_torque} shows that, qualitatively, the inverse dynamics model matches the measurement of the motor torques.

Torque margin is defined in equation \eqref{eq:torque_margin}, since motors have a speed-torque curve due to their generated Back EMF and limited inverter bus voltage, torque margin is, in general, a function of the output load torque/thrust $\tau$ and joint output speed $\omega$. For gearboxes, the maximum torque rating $\tau_{max}$ is typically independent of gearbox speed so this dependence vanishes.

\begin{equation} \label{eq:torque_margin}
    \tau_{margin}(\tau,\ \omega) = (1-|\tau/\tau_{max}(\omega)|)\times100\%
\end{equation}

\begin{figure}[!t]
\vspace{6pt} 
    \centering
    \includegraphics[width=8cm]{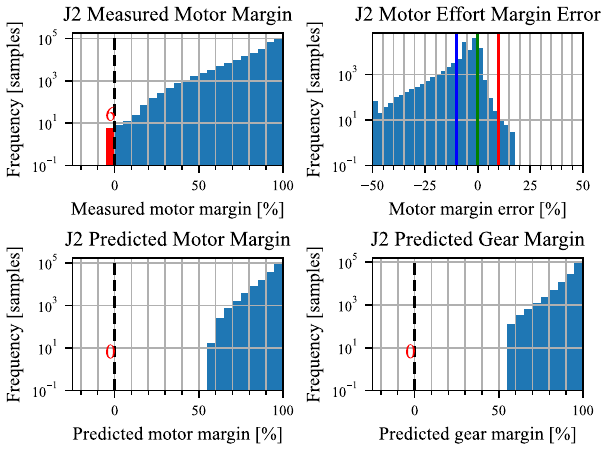}
    \caption{\textbf{Histogram of J2 motor and J2 gearbox torque margins over the course of a match with an elite player before the motor upgrade.} Y axes are logarithmic, torque margin data is sampled at 1 kHz. Top-left: Cumulative histogram of motor margins from the measured motor torque. Top-right: Histogram of motor torque margin errors. Bottom-left: Cumulative histogram of predicted motor torque margins from the inverse dynamics model. Bottom-right: Cumulative histogram of predicted gearbox torque margins from the inverse dynamics model.}
    \label{fig:j2_histogram_20251202}
\end{figure}

\begin{figure}[!t]
    \centering
    \includegraphics[width=8.5cm]{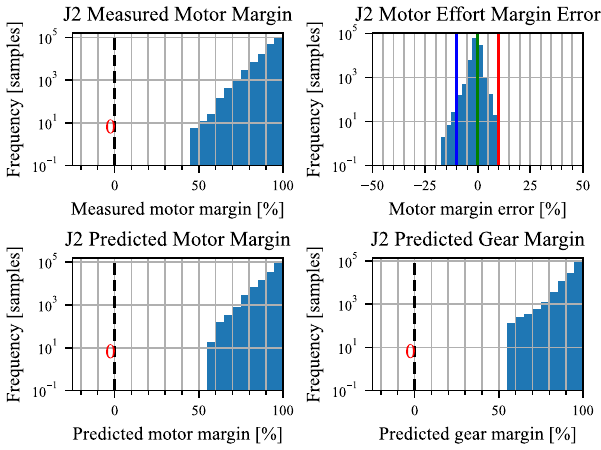}
    \caption{\textbf{Histogram of J2 motor and J2 gearbox torque margins over the course of a match with an elite player after the motor upgrade.} Y axes are logarithmic, torque margin data is sampled at 1 kHz. Top-left: Cumulative histogram of motor margins from the measured motor torque. Top-right: Histogram of motor torque margin errors. Bottom-left: Cumulative histogram of predicted motor torque margins from the inverse dynamics model. Bottom-right: Cumulative histogram of predicted gearbox torque margins from the inverse dynamics model.}
    \label{fig:j2_histogram_20260210}
\end{figure}

\begin{table}[!t]
\vspace{6pt} 
\centering
\caption{minimum torque margins for each joint for two actuator configurations}
\label{tab:torque_margins}
\scalebox{0.9}{
\begin{tabular}{lccc}
\hline
Component & Baseline (Torque margin \%)  & Revised (Torque margin \%) \\
\hline
X Motor         & 48.09 & 47.96 \\
Y Motor         & 56.33 & 56.14 \\
J1 Motor        & 26.42 & 38.27 \\
J1 Gearbox      & 68.71 & 70.2 \\
J2 Motor        & 5.65 & 53.53 \\
J2 Gearbox      & 38.45 & 41.44 \\
J3 Motor        & 6.49 & 11.31 \\
J3 Gearbox      & 8.57 & 14.0 \\
J4 Motor        & 9.48 & 16.03 \\
J4 Gearbox      & 5.88 & 13.48 \\
J5 Motor        & 44.07 & 46.52 \\
J5 Gearbox      & 50.53 & 52.79 \\
J6 Motor        & 70.92 & 73.93 \\
J6 Gearbox      & 95.43 & 83.23 \\
\hline
\end{tabular}}
\end{table}

For the optimization, a dataset of robot joint trajectories was prepared from data collected during multiple games with elite table tennis players and used as a reference design trajectory for the joint re-dimensioning analysis. The \emph{minimum} torque margins predicted during these motions were computed for different actuator configurations using the inverse dynamics model. The various configurations were compared and the configuration offering the highest minimum margin was selected. The minimum torque margins experienced during a defined motion for each joint and for two such actuator configurations are shown in Table \ref{tab:torque_margins}. The first column represents a baseline corresponding to an initial design configuration of the robot while the second column represents a revised design where the J1 and J2 motors are upgraded one catalog size up and weight is reduced from the J6 transmission by replacing steel gears with plastic ones. This analysis motivated upgrading the J1 and J2 motors but the switch to plastic gears was foregone due to durability concerns.

From this analysis it was determined that it would be beneficial to increase the size of the J1 and J2 motors to the next available size from the manufacturer's product line. 

Fig.~\ref{fig:j2_histogram_20251202} shows histograms of the gearbox and torque margins before the J2 motor upgrade and Fig.~\ref{fig:j2_histogram_20260210} shows the same histograms after the motor upgrade. It is clear that motor torque margins are significantly increased as a result.

%%%%%%%%%%%%%%%%%%%%%%%%%%%%%%%%%%%%%%%%%%%%%%%%%%%%%%%%%%%%%%%%%%%%%%%%%%%%%%%%%%%%%%%%
\section{Robot dynamics model identification}
%%%%%%%%%%%%%%%%%%%%%%%%%%%%%%%%%%%%%%%%%%%%%%%%%%%%%%%%%%%%%%%%%%%%%%%%%%%%%%%%%%%%%%%%
\begin{table*}[!t]
\vspace{6pt} 
\caption{Identified parameters of the process models for all joints with uncertainties}
\label{tab:process_model_params}
\centering
\renewcommand{\arraystretch}{1.3}
\scalebox{0.95}{
\begin{tabular}{l|llllllll}
                  & X                            & Y                           & J1                           & J2                          & J3                          & J4                          & J5                & J6                 \\ \hline
$\tau_d~[\si{s}]\times 10^{-3}$ & $1$              & $1$             & $2$              & $1$             & $0$                       & $2$             & $1$             & $3$               \\
$T_w~[\si{s}]\times 10^{-2}$    & $2.13\pm0.05$  & $2.84\pm0.23$ & $1.92\pm0.03$  & $0.15\pm1.71$  & $29.1\pm8.6$ & $1.21\pm0.02$ & $0.17\pm0.09$ & $0.24\pm0.00$ \\
$T_z~[\si{s}]\times 10^{-1}$    & $0.95\pm0.04$  & $1.80\pm0.29$ & $1.55\pm0.05$  & $0.00\pm0.18$               & $195\pm116$              & $0.60\pm0.02$               & $0.00\pm0.01$    & $0.04\pm0.00$  \\
$\zeta$           & $2.33\pm0.05$                & $3.25\pm0.25$               & $4.10\pm0.07$                & $1.00\pm5.52$                & $33.6\pm10.0$              & $2.60\pm0.04$               & $1.00\pm0.25$    & $1.06\pm0.01$    
\end{tabular}
}
\end{table*}

\begin{table*}[!t]
\caption{RMSE and maximum errors of the dynamics model integration}
\label{table:sysid}
\centering
\renewcommand{\arraystretch}{1.3}
\scalebox{0.95}{
\begin{tabular}{l|llllllll}
     & X $[\si{m}]$      & Y $[\si{m}]$      & J1 $[\si{rad}]$    & J2 $[\si{rad}]$     & J3 $[\si{rad}]$    & J4 $[\si{rad}]$     & J5 $[\si{rad}]$    & J6 $[\si{rad}]$    \\  \hline
RMSE & $1.6\times10^{-5}$ & $1.1\times10^{-5}$ & $2.1\times10^{-5}$ & $1.0\times10^{-5}$ & $8.2\times10^{-4}$ & $1.7\times10^{-5}$ & $8.1\times10^{-6}$ & $1.8\times10^{-5}$ \\
Max  & $9.4\times10^{-5}$  & $6.8\times10^{-5}$ & $2.2\times10^{-4}$ & $8.2\times10^{-5}$ & $8.7\times10^{-3}$ & $1.6\times10^{-4}$  & $6.1\times10^{-5}$ & $2.1\times10^{-4}$ 
\end{tabular}
}
\end{table*}
In order to develop RL control policies for \ace{} 
in simulation, we needed an accurate representation of the dynamics of \ace{}.
We identified a per-joint model for \ace{} 
using Matlab\cite{sysid_toolbox_matlab}, assuming independent joint dynamics. 
Given the utilization of reduction gears ($80:1$ for J1, J2 and J5, $57.5:1$ for J4, $50:1$ for J3, $20:1$ for J6) and the proper tuning of high-gain servo systems for each axis, inter-joint coupling effects are sufficiently attenuated at the actuator side, making this approximation reasonable within the considered operating conditions.
For every joint, we used a two-stage gray-box workflow in which a flexible model is first identified for initialization, and then refined into a low-order process form. 
This method also assumes a delayed single-input single-output mapping from commanded position to measured position, and the final result of the pipeline is a continuous-time dynamics model with fixed order, i.e. state vector size, of $2$.
This is a hyperparameter of our identification method, that was determined heuristically as the smallest value after which further increasing it does not provide a significant improvement in the model prediction error. 
The state $\bm{x}$ of every joint is thus described by its position, and an additional sub-state for improved numerical modeling.

The initial model was obtained via continuous-time subspace identification with process disturbance and fixed input delay $\tau_d$, defined as:
\begin{align}
    \label{eq:ss_model}
    \dot{\bm{x}}(t)&=\bm{A}\bm{x}(t)+\bm{B}\bm{u}(t-\tau_d)+\bm{K}\bm{e}(t) \\
    y(t)&=\bm{C}\bm{x}(t)+\bm{e}(t),
\end{align}
with $\bm{C}$ corresponding to the matrix $[1, 0]$, and $\bm{e}(t)\sim\mathcal{N}(\bm{0}, \bm{\Lambda})$ describing the identified system disturbance, sampled from a zero-mean Normal distribution, and $\bm{K}$ is the estimated gain matrix.
Then, we converted the obtained model to a second-order process model with lead time and delay, generally described as in \eqref{eq:underdamped_model}.
\begin{align}
    \label{eq:underdamped_model}
    \bm{G}(\bm{s}) = K_p \frac{1+T_z \bm{s}}{T_w^2 \bm{s}^2 + 2\zeta T_w \bm{s} + 1}e^{-\tau_d \bm{s}}
\end{align}
We optimized the values for the natural period, zero time, and damping parameters $(T_w, T_z, \zeta)$ based on the initial values calculated from the zeros and poles of the state space model obtained during the previous step.
The noise-to-output transfer function from the state-space model was also extracted and used to weight the residuals during the optimization, to model colored noise.
The steady-state gain was fixed to $K_p=1$, and the time delay $\tau_d$ was kept constant, but a binary search was conducted for every joint to find the value of $\tau_d$ that minimized the RMSE.
We used this search approach for $\tau_d$ since we wanted to have it be a multiple of $1~\si{ms}$ (the control period of \ace{}
), in order to make the model more flexible to use.
The obtained parameters for all the joints are summarized in Table \ref{tab:process_model_params}.
Looking at the values found, we confirmed that J2, J5 and J6 have damping coefficients close to 1 (i.e. almost critically damped) and small values for $T_z$ and $T_w$, which is the most desirable scenario and indicates that these joints in particular are the least affected by mechanical resonance.
On the other hand, J3 was the opposite case, often experiencing high disturbances which made the low-level control tuning more challenging, and produced a more under-damped system, and higher uncertainty in the identified coefficients. 

Finally, the optimized models were converted back to a state space model and evaluated on a validation dataset containing an input and measured position trajectory pair sampled at $1~\si{kHz}$.
This data represented a concatenated collection of table tennis strikes recorded during an experimental session against a real player, amounting to about 62 seconds of nearly uninterrupted motion at a variety of speeds and accelerations.
The prediction errors obtained are summarized in Table \ref{table:sysid}.
We report the RMSE and maximum errors of the predicted joint states calculated by integrating the identified models over the full input trajectory, compared against the measured joint positions.
The results were consistent with our analysis of Table \ref{tab:process_model_params}, where the errors of J2, J5, J6 are the lowest and J3 the highest.
Assuming the obtained average model error over 1000 random robot configurations, we estimated the average positional and rotational error at the racket frame, which resulted in $0.056~\si{cm}$ and $9\times10^{-4}~\si{rad}$ respectively, well within our target specifications in Table \ref{tab:target_spec}.

%%%%%%%%%%%%%%%%%%%%%%%%%%%%%%%%%%%%%%%%%%%%%%%%%%%%%%%%%%%%%%%%%%%%%%%%%%%%%%%%%%%%%%%%
\section{Control System Hardware}
%%%%%%%%%%%%%%%%%%%%%%%%%%%%%%%%%%%%%%%%%%%%%%%%%%%%%%%%%%%%%%%%%%%%%%%%%%%%%%%%%%%%%%%%
\begin{figure}[!t]
\centering
\includegraphics[width=8.5cm]{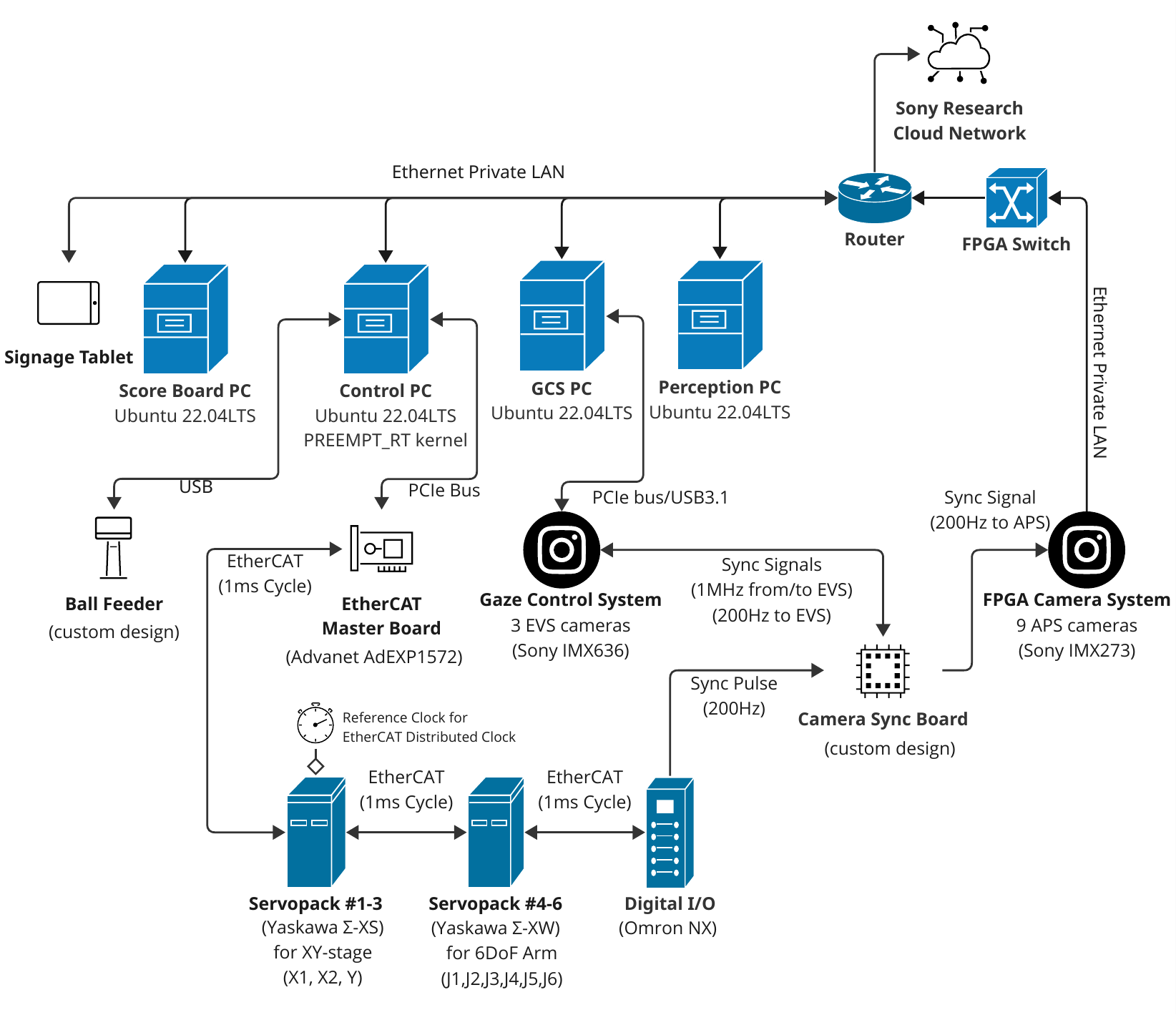}
\caption{\textbf{Hardware configuration of the control system.} 
The Control PC generates target position trajectories, drives actuators by sending commands to the servo amplifiers acting as EtherCAT slaves via an EtherCAT master board, synchronizes the camera system, and controls the ball feeder and signage tablet.
The GCS PC measures ball spin with the gaze control system.
The Perception PC measures ball position with the FPGA camera system.
Scoreboard PC calculates and displays scores and orchestrates logging data.
}
\label{fig:control system}
\end{figure}

Fig.~\ref{fig:control system}  shows the hardware configuration of the control system. Based on information obtained from the Perception PC, which measures ball position by using APS (Active-Pixel Sensor) cameras, and the GCS (Gaze Control System) PC, which measures ball spin by using EVS (Event-based Vision Sensor) cameras,
\cite{Hu2026GCS} , 
the Control PC generates target position trajectories for the three linear actuators and the six rotary actuators of \ace{}. 
Note that two linear actuators are used in parallel to drive the X-axis of the XY-stage.
The Control PC also controls a custom-designed ball feeder, placing a ball into the ball cup when serving, and displays information regarding the robot's status (e.g., preparing, ready to play) to human opponents.
Lastly, the Scoreboard PC calculates and displays scores and orchestrates logging data.

The generated target trajectories are commanded to the nine actuators by transmitting  target positions to each Yaskawa Sigma-X servo amplifier \cite{Yaskawa} at $1~\si{ms}$ intervals using CSP (Cyclic Synchronous Position) mode over EtherCAT (Ethernet for Control Automation Technology) \cite{EtherCAT} from an Advanet AdEXP1572 EtherCAT master board \cite{advanet} installed in the Control PC.

By utilizing the distributed clock function of EtherCAT, all EtherCAT slaves are synchronized within $1~\si{\micro s}$. 
A $200~\si{Hz}$ pulse, synchronized with the EtherCAT cycle and output from the digital I/O unit, facilitates the synchronization of three EVS cameras with Sony IMX636 sensors for the GCS PC, nine APS cameras with Sony IMX273 sensors for the Perception PC, and the nine actuators of \ace{}.

%%%%%%%%%%%%%%%%%%%%%%%%%%%%%%%%%%%%%%%%%%%%%%%%%%%%%%%%%%%%%%%%%%%%%%%%%%%%%%%%%%%%%%%%
\section{Experimental Results}
%%%%%%%%%%%%%%%%%%%%%%%%%%%%%%%%%%%%%%%%%%%%%%%%%%%%%%%%%%%%%%%%%%%%%%%%%%%%%%%%%%%%%%%%

\begin{figure}[!t]
\vspace{6pt} 
\centering
\includegraphics[width=8.5cm]{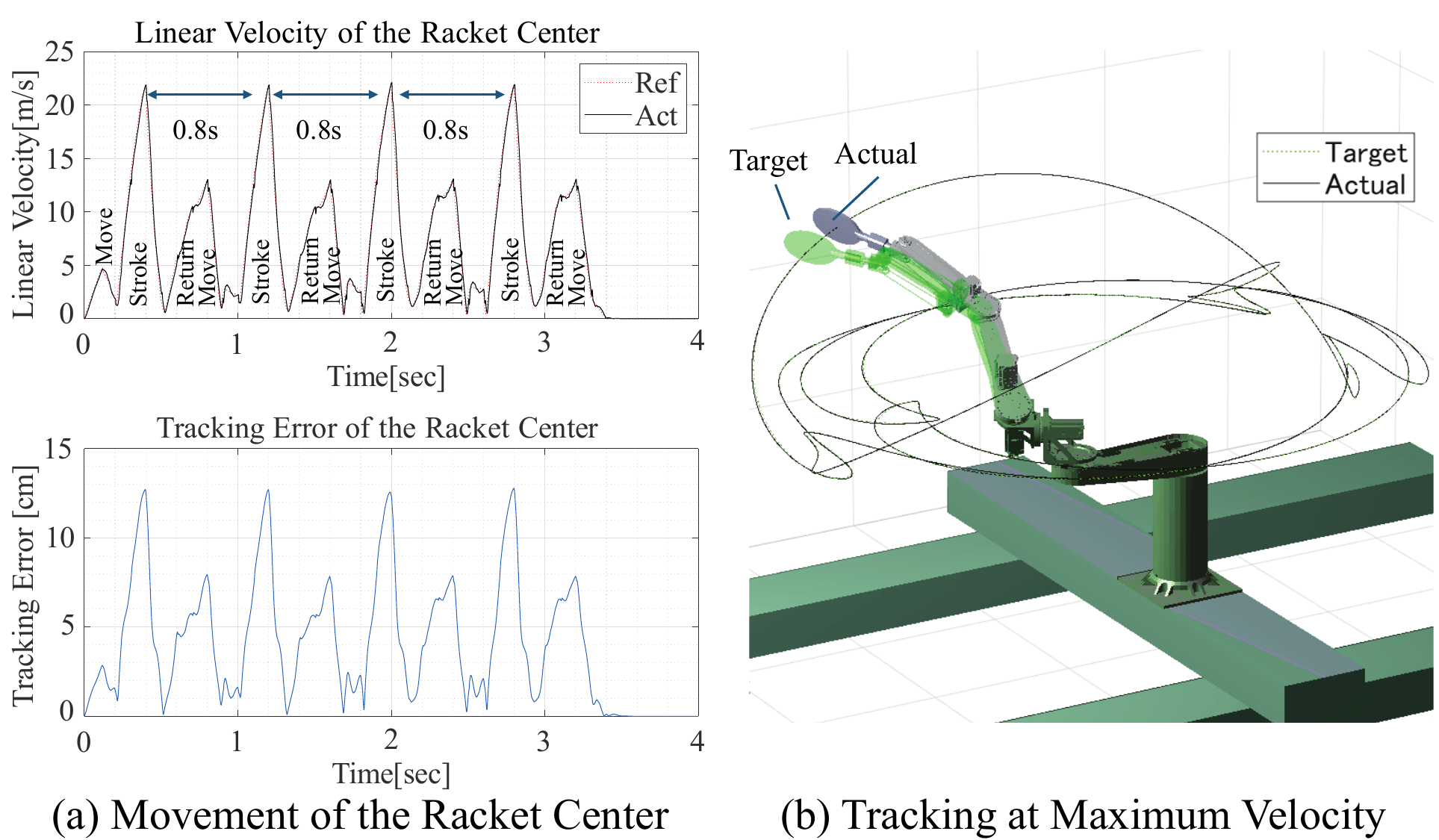}
\caption{\textbf{Full-stroke swing experiment at $0.8~\si{s}$  intervals.}
(a) Linear velocity and tracking error of the center of the racket during four round-trip swings.  
(b) 3D visualization of the target and actual trajectories of the center of the racket and the posture of the arm at maximum velocity.
}
\label{fig:swing}
\end{figure}

\begin{figure}[!t]
\vspace{6pt} 
\centering
\includegraphics[width=8.5cm]{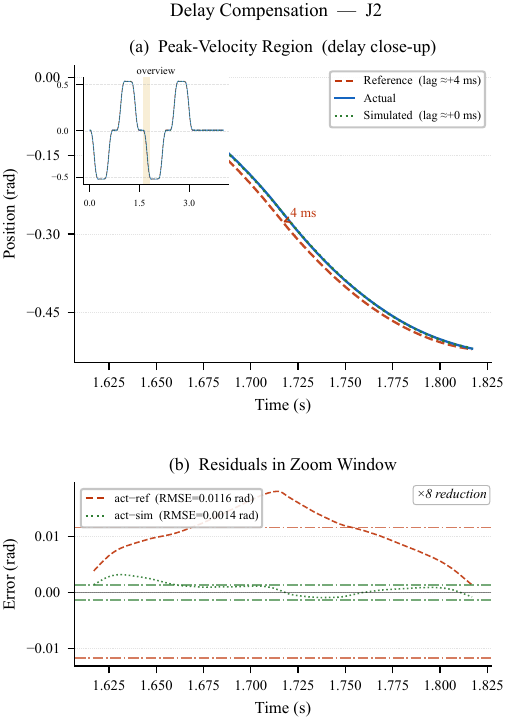}
\caption{\textbf{Delay compensation in the peak-velocity region for J2.} (a) Close-up of the highest-acceleration segment of the trajectory, where the tracking delay is most visible. The reference trajectory (red dashed) is visibly phase-shifted relative to the actual motion (blue solid) by $\approx +4ms$, while the model prediction (green dotted) tracks the actual position with negligible lag. The inset shows the full trajectory with the zoomed region highlighted. (b) Tracking error (act − ref, red dashed) and model residual (act − sim, green dotted) within the same window; dashed-dotted lines mark $\pm$RMSE for each. 
%The model reduces the residual by over an order of magnitude compared to the open-loop command.
The model residuals are minimal at 12 \% of the original tracking error.
}
\label{fig:fig5b_delay_compact}
\end{figure}

\begin{figure}[!t]
\vspace{6pt} 
\centering
\includegraphics[width=8cm]{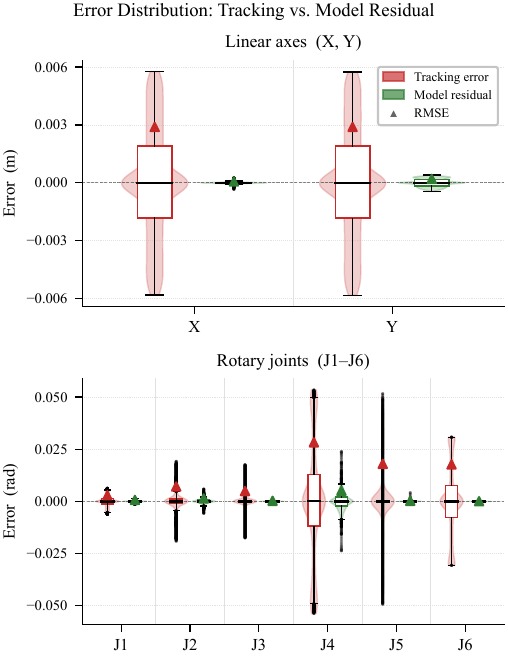}
\caption{\textbf{Tracking error statistics per joint.} For each joint, two side-by-side violin–box plots compare the tracking error (act − ref, red) and the model residual (act − sim, green) over the full trajectory. Violin shapes indicate the empirical distribution; box plots show the interquartile range and median. Linear axes (X, Y) and rotary joints (J1–J6) are shown in separate panels.}
\label{fig:fig1c_error_statistics}
\end{figure}

Fig.~\ref{fig:swing} shows the results of \ace{} 
repeatedly performing a full-stroke swing while moving $\pm0.3~\si{m}$ in the X and Y directions, with a cycle time of $0.8~\si{s}$ as the design target (see the supplementary video).
These results show that a maximum velocity of $22~\si{m/s}$ can be achieved even at a $0.8~\si{s}$ cycle, with almost no deviation from the trajectory at the motor encoder level. 
This indicates that the targeted physical performance has been achieved in the robot hardware design.

The tracking error is approximately proportional to the linear velocity, peaking at $12.8~\si{cm}$ at maximum linear velocity.
This suggests that the tracking delay of the low-level control system, which includes communication latency, is the dominant factor of the tracking error, and that this delay is less than $6~\si{ms}$.

Fig.~\ref{fig:fig5b_delay_compact} and Fig.~\ref{fig:fig1c_error_statistics} illustrate the effect of delay compensation on the trajectory-tracking residuals. In the peak-velocity region, where delay effects are most visible, Fig.~\ref{fig:fig5b_delay_compact}a shows a clear temporal offset between the commanded reference and the measured motion for one joint. In contrast, the delayed model prediction remains closely aligned with the actual trajectory, indicating that the identified delay captures the dominant phase mismatch in this dynamic portion of the motion. This is further supported by Fig.~\ref{fig:fig5b_delay_compact}b, where the model residual, defined as act$-$sim, is substantially smaller than the raw tracking error, act$-$ref, over the same interval. The full-trajectory statistics in Fig.~\ref{fig:fig1c_error_statistics} confirm that this improvement is not limited to a local window: across both the linear axes and rotary joints, the residual distributions are consistently narrower and more concentrated around zero than the corresponding tracking-error distributions. Overall, these results show that the exploited delay-compensated model explains a significant portion of the tracking error across the robot trajectory.
The use of this delay-compensated model in simulations enables the training of highly capable RL control policies without the necessity of training on actual robot hardware.

%%%%%%%%%%%%%%%%%%%%%%%%%%%%%%%%%%%%%%%%%%%%%%%%%%%%%%%%%%%%%%%%%%%%%%%%%%%%%%%%%%%%%%%%
\section{Conclusions}
%%%%%%%%%%%%%%%%%%%%%%%%%%%%%%%%%%%%%%%%%%%%%%%%%%%%%%%%%%%%%%%%%%%%%%%%%%%%%%%%%%%%%%%%

For the first time, the hardware design specifications required for a table tennis robot to compete against professional players under official competition rules have been elucidated.
Based on these specifications, a robot and its control system were developed. We identified delay-compensated dynamic models with sufficient accuracy and predictability. The RL control policies that were trained through simulations using these models were directly applicable to matches against human players, and were proven capable of defeating professional players. 
This not only validated the design specifications but also exemplified that an autonomous robot is able to outperformed professional experts in real-world tasks, provided the hardware possessed sufficient performance.

Miu Hirano, an Olympic silver medalist who played against \ace{} commented that it would rank in the top 10 among women and around 50th among men.
We posit that with current hardware, if match data is accumulated and RL control policies are refined, \ace{} 
could potentially defeat the world's top-ranked male player, but the adaptability of high-ranking players may challenge this. 
If the hardware performance of sports robots far exceeds that of humans,  
the excitement of the matches would be lost.
Consequently, the present hardware performance of \ace{} 
could be regarded as a good balance for table tennis robot design.

%%%%%%%%%%%%%%%%%%%%%%%%%%%%%%%%%%%%%%%%%%%%%%%%%%%%%%%%%%%%%%%%%%%%%%%%%%%%%%%%%%%%%%%%
\section*{Acknowledgments}
%%%%%%%%%%%%%%%%%%%%%%%%%%%%%%%%%%%%%%%%%%%%%%%%%%%%%%%%%%%%%%%%%%%%%%%%%%%%%%%%%%%%%%%%
We thank all the professional players who played with \ace{} and provided feedback: Miu Hirano, Takeru Sone, Seiya Kishikawa, Miyuu Kihara, Takako Nagao, Maki Shiomi, Minami Ando, Mayuka Taira, Kazuhiro Yoshimura, Daito Ono, Shunsuke Okano, Tonin Ryuzaki, Fumiya Igarashi; as well as our coach, Yamato Kawamata, and the university league players.
We also thank Koji Matsushita and all the staff at VICTAS for their advice and support.
In addition, we thank all 
%We also thank Mireille El Gheche, Guilherme Jorge Maeda, Naoya Takahashi, Hamdi Sahloul, Yamen Saraiji, Yin Bi, Sam Blakeman, Christian Conti, Dunai Fuentes Hitos, Yunpu Hu, Raphaela Kreiser, Luz Martinez, Fabian Schilling, Ricardo Tapiador Morales, Mario Ynocente Castro, Lison Abecassis, Alberto Giammarino, Yu-Ting Huang, Yannik Nagel, Andrea Scotti, Tiago Silva, Etienne Walther, Jengyan Wong, Bilan Yang, Asude Aydin, Apurv Saha, Megumu Tsukamoto, Taishi Kunori, Valentin Monferrato, 
%and other 
ACE project teammates for data collection for robot development, support for daily experiments, and achieving outstanding results made using the developed robots;
as well as Stefan Richter for his earlier help with system identification. 
Finally, we thank Peter Stone and Michael Spranger for their advice and support, and Hiroaki Kitano for giving us the opportunity to develop this research.

\bibliographystyle{IEEEtran}
\bibliography{IEEEabrv,aceHW}

\vfill

\end{document}